\documentclass[runningheads]{llncs}

\usepackage{accv}

\usepackage{accvabbrv}

\usepackage{graphicx}
\usepackage{booktabs}
\usepackage{multirow}

\usepackage[accsupp]{axessibility}  % Improves PDF readability for those with disabilities.

\usepackage{hyperref}

\usepackage{orcidlink}
\usepackage{xcolor}
\usepackage[most]{tcolorbox}

\definecolor{soccerGreen}{RGB}{46, 125, 50}
\newtcolorbox{promptbox}[1][]{
    colback=soccerGreen!6,
    colframe=soccerGreen!45,
    colbacktitle=soccerGreen!18,
    coltitle=black,
    fonttitle=\bfseries\small,
    title={#1},
    boxrule=0.5pt,
    arc=2mm,
    left=2mm,
    right=2mm,
    top=1.5mm,
    bottom=1.5mm
}

\begin{document}
\raggedbottom

\title{SoccerNet-FoulRet: Retrieving Semantically Similar Soccer Foul Videos} 

\titlerunning{SoccerNet-FoulRet: Retrieving Semantically Similar Soccer Foul Videos}

\author{
Jacobus Arthur\inst{1}\textsuperscript{*} \and
Ahmad Sait\inst{2}\textsuperscript{*} \and
Batool Hani\inst{2} \and
Merey Ramazanova\inst{2} \and
Jan Held\inst{3} \and
Marc Van Droogenbroeck\inst{1} \and
Bernard Ghanem\inst{2} \and
Anthony Cioppa\inst{1} \and
Silvio Giancola\inst{2}
}

\authorrunning{J.~Arthur, A.~Sait et al.}

\institute{University of Liège
\email{\{anthony.cioppa, m.vandroogenbroeck\}@uliege.be, arthur.jacobus@student.uliege.be}\\ \and
KAUST
\email{\{ahmed.sait, batool.hani, merey.ramazanova, bernard.ghanem, silvio.giancola\}@kaust.edu.sa}\\ \and
SpAItial
\email{jan@spatial.ai}}

\maketitle
\begin{center}
\small\textsuperscript{*}Equal contribution.
\end{center}

\begin{abstract}
  Refereeing decisions in professional soccer remain inconsistent because referees cannot
  easily compare a contentious foul against similar past cases. We cast this as a retrieval
  problem and introduce \textbf{SoccerNet-FoulRet}, the first benchmark for \emph{semantic}
  foul retrieval. Given a query foul, the task is to retrieve past fouls judged to be relevant precedents, regardless of camera angle, teams, or appearance. This differs from prior
  video-to-video retrieval, which matches clips by visual similarity or a shared event. Here,
  relevance is defined by refereeing interpretation. We build the benchmark from the SoccerNet-MVFoul dataset and evaluate retrieval ability of zero-shot video and vision--language embedders together with a task-specific fine-tuned baseline on 693 human-verified queries and category-relevance labels. Semantic foul retrieval remains challenging. The strongest zero-shot model achieves under $5\%$ HitRate@10 on human-verified precedents, while category-supervised fine-tuning improves category relevance but transfers only modestly to precedent retrieval. We release SoccerNet-FoulRet to establish semantic foul retrieval as an open problem: \href{https://github.com/SoccerNet/sn-foulret}{https://github.com/SoccerNet/sn-foulret}.
  
\end{abstract}

\section{Introduction}
\label{sec:intro}

\begin{figure}[t]
\centering
\includegraphics[width=\linewidth]{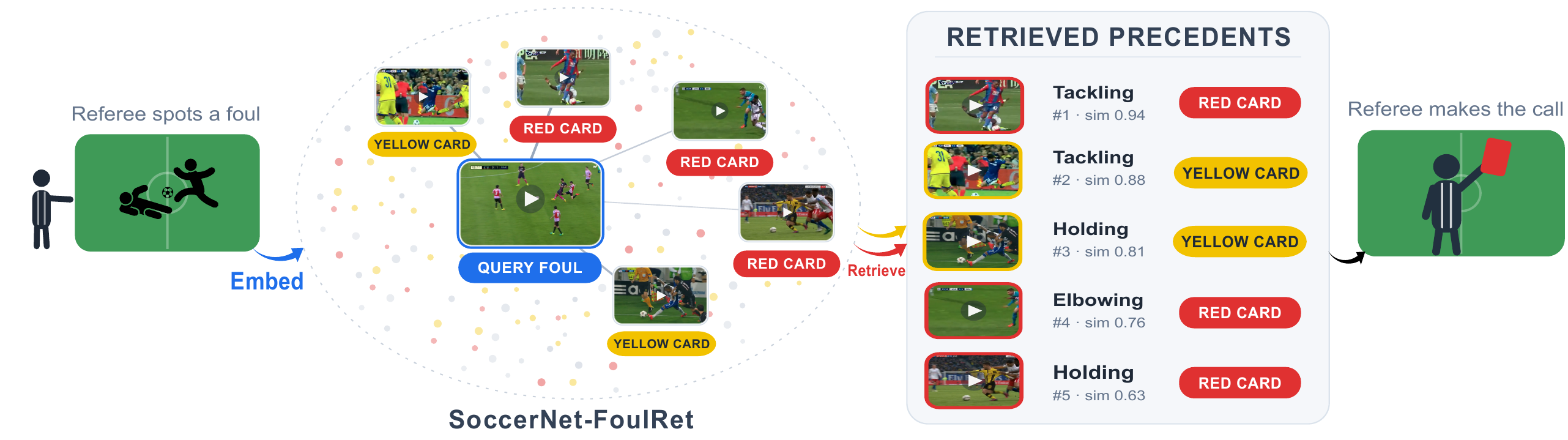}
\caption{\textbf{A memory bank for referees.} We aim to \emph{augment}
referees rather than replace them. Automated classifiers and generative explainers emit a
verdict directly, but can do so overconfidently and even hallucinate. Instead, our system takes a new foul (left) and retrieves semantically similar precedents together with the decisions that past referees reached on them (right). The referee reasons from these precedents and keeps the final call. We introduce a human-verified benchmark for this video-to-video retrieval task and find that off-the-shelf zero-shot video and vision-language retrievers fall well short, leaving it an open problem.}
\label{fig:teaser}
\end{figure}

The modernization of soccer arbitration, notably through video assistant referee (VAR) technology, has provided referees with better visual evidence to judge complex situations. However, the interpretation of this evidence remains inherently subjective. Two referees may judge the same contact differently based on personal criteria or viewing angles.
As a consequence, similar actions may be judged differently depending on the referee, leading to inconsistent decisions that vary from one match to another.
The major limitation of the current decision process is the isolation of the decision process. Referees cannot easily access or compare the current incident with similar historical cases to ensure consistency.

In this work, we develop a retrieval tool designed to assist referees by finding similar fouls (\cref{fig:teaser}). Developing such a tool presents a dual challenge. First, from a methodological standpoint, the system must translate raw video pixels into high-level semantic concepts (\eg, ``intent'', ``severity'') to perform accurate retrieval. Second, from an experimental standpoint, there is currently no standard benchmark or ground truth for foul similarity, making it difficult to measure the effectiveness of such a tool.

We address these challenges by contributing a benchmark and evaluation protocol for
foul retrieval, together with a study of how existing methods perform on it.
Our central contribution is \textbf{SoccerNet-FoulRet}, the first benchmark for semantic
foul retrieval. Since the source dataset, SoccerNet-MVFoul, provides no similarity links between foul incidents, we construct them ourselves. We build a search gallery by merging existing ground-truth annotations and develop a manual validation interface to construct a \textit{stare decisis} precedent base, which is a collection of query--candidate pairs with human-verified precedent judgments. This allows us to evaluate retrieval against human judgments of precedent relevance.

On this benchmark, we evaluate a range of retrieval approaches, including a
text-mediated VARS+XVARS pipeline, zero-shot video and vision--language
embedders, and a task-specific fine-tuned baseline. This comparison tests
whether current retrieval methods can retrieve human-verified foul precedents
and whether category-level supervision transfers to precedent relevance.
Framing refereeing consistency as a retrieval problem allows officials to
compare new incidents with similar historical fouls, supporting more consistent
decision-making.

We summarize our contributions as follows.
\textbf{(i)} We introduce the first benchmark for soccer foul retrieval, built from
human-verified judgments of precedent relevance.
\textbf{(ii)} We benchmark a text-mediated retrieval pipeline, zero-shot video and
vision--language embedders, and a task-specific fine-tuned baseline, evaluating both
human-verified precedent retrieval and category-level relevance.
\textbf{(iii)} We show that category-level supervision improves retrieval by action,
offence, and severity, but transfers only modestly to human-verified precedent
relevance, highlighting a gap between categorical similarity and human judgments
of precedent relevance.

\section{Related Work}
\label{sec:related_work}
We situate our work within two lines of research: sports video understanding and
video-to-video retrieval.

\subsection{Sports Video Understanding}
Automated sports analysis has evolved from simple classification tasks to complex event understanding, driven by large-scale benchmarks spanning action spotting~\cite{giancola2018soccernet, deliege2021soccernetv2}, player tracking and re-identification~\cite{Cioppa2022Scaling, Cioppa2022SoccerNetTracking, cui2023sportsmot}, and more recently language-driven understanding such as dense captioning~\cite{Mkhallati2023SoccerNetCaption} and game state reconstruction~\cite{Somers2024SoccerNetGameState}.

Despite this breadth, refereeing assistance remains under-explored. Held
\etal~\cite{Held2023VARS} introduced SoccerNet-MVFoul, the first multi-view dataset
dedicated to foul classification, and SoccerNet-XFoul~\cite{Held2024XVARS} extended it with
explainable question-answer pairs. Our work builds on both. We use SoccerNet-MVFoul's
structured foul annotations and SoccerNet-XFoul's referee-oriented explanations to generate
candidate pairs for human validation, while the MVFoul labels additionally support our
category-level evaluation. Prior work uses SoccerNet-MVFoul for foul
classification~\cite{Held2023VARS} and SoccerNet-XFoul for video description, question
answering, and action recognition~\cite{Held2024XVARS}, whereas we repurpose them for
retrieval. This enables the study of a different problem: retrieving historical fouls that
serve as relevant precedents.

\subsection{Video-to-Video Retrieval}

Video-to-video retrieval has largely been framed around visual overlap, from
near-duplicate copies \cite{Wu2007Practical, Song2011Multiple, Jiang2019SVD} to shared real-world events. \Cref{tab:datasets} provides a structured comparison of existing benchmarks alongside our proposed dataset.

\noindent\textbf{Near-Duplicate Video Retrieval (NDVR).} NDVR is the task of
finding visually redundant copies of a query video within a large gallery,
typically web videos that have been re-uploaded, re-encoded, or slightly edited. Benchmarks such as CC\_WEB\_VIDEO~\cite{Wu2007Practical}, UQ\_VIDEO~\cite{Song2011Multiple}, and SVD~\cite{Jiang2019SVD} have driven significant progress in this area. However, they are fundamentally ill-suited to our setting. Two soccer fouls may be visually near-identical yet jurisprudentially distinct, and conversely, two fouls judged to form a relevant precedent pair may appear visually dissimilar due to differences in camera angle or player appearance.

\noindent\textbf{Video Copy Detection.} Copy detection targets a stricter
variant of NDVR. It identifies whether a segment of one video is derived from
another, possibly after transformations such as cropping, re-encoding, color
changes, or frame-rate modifications~\cite{Kraaij2011TRECVID}. The query and its ground-truth match necessarily originate from the same source recording. VCDB~\cite{Jiang2014VCDB} provides manually annotated copied segment pairs from web videos, while TRECVID-CBCD~\cite{Kraaij2011TRECVID} relies on synthetically generated queries (\ie, artificially transformed duplicates), making the task less reflective of naturally
occurring video copies. Neither paradigm applies to our setting, where two similar fouls occur in entirely independent matches with no shared source footage.

\noindent\textbf{Incident and Event Retrieval.} These benchmarks move
beyond visual redundancy by grouping videos around shared real-world incidents. FIVR-200K~\cite{KordopatisZilos2019FIVR} retrieves videos depicting the same news event from different sources (\eg, multiple recordings of the same accident or protest), while EVVE~\cite{Revaud2013Event} targets the same principle at the scale of large broadcast archives. Although this represents a step toward semantic retrieval, similarity remains anchored to a shared physical event. In our setting, no such shared event exists. Two fouls from different matches may instead form a relevant precedent pair based on human judgment, a criterion these benchmarks do not capture.

\noindent\textbf{Concept-Based Semantic Retrieval.} Closest to our setting,
ConViS-Bench~\cite{Liberatori2025ConViSBench} ranks videos by shared semantic concepts rather than
visual redundancy, and reports that this remains difficult for current large multimodal
models. Its relevance is still defined by general-purpose concepts such as location or
action, and it does not target soccer.

Across these settings, retrieval systems encode each video into a vector and rank
by embedding distance. Near-duplicate and copy-detection methods train task-specific visual
embeddings, whereas recent multimodal embedders learn general-purpose video-text spaces
that support zero-shot retrieval. We evaluate several such zero-shot models together with
a task-specific adaptation baseline (\cref{sec:experiments}). Semantic
relevance itself has also been studied in text-to-video retrieval, for example through caption correspondence~\cite{Wray2021OnSemantic}. We instead evaluate human-verified precedent relevance, where query–-candidate pairs are judged as relevant or irrelevant precedents. No prior benchmark captures this human-judgment criterion.

\begin{table}[t]
\centering\small\setlength{\tabcolsep}{4pt}
\resizebox{\columnwidth}{!}{%
\begin{tabular}{l c c c c l}
\toprule
Dataset & \#Videos & \#Queries & Query Origin & Domain & Task \\
\midrule
CC\_WEB\_VIDEO~\cite{Wu2007Practical} & 13K   & 24    & Real      & Web (YouTube) & NDVR \\
UQ\_VIDEO~\cite{Song2011Multiple}        & 170K  & 24    & Real      & Web (YouTube) & NDVR \\
SVD~\cite{Jiang2019SVD}                  & 562K  & 1,206 & Real      & Web (Douyin)  & NDVR \\
VCDB~\cite{Jiang2014VCDB}                & 100K  & 528   & Real      & Web (YouTube) & Copy Detection \\
TRECVID-CBCD~\cite{Kraaij2011TRECVID}     & 11,503 & 11,256 & Synthetic & Broadcast TV  & Copy Detection \\
FIVR-200K~\cite{KordopatisZilos2019FIVR}           & 226K  & 100   & Real      & Web (YouTube) & Fine-grained IR \\
EVVE~\cite{Revaud2013Event}                & 102K  & 620   & Real      & Web (YouTube) & Event Retrieval \\
ConViS-Bench~\cite{Liberatori2025ConViSBench} & 543   & 610   & Real      & Web (general) & Concept Similarity \\
\midrule
\textbf{SoccerNet-FoulRet (Ours)} & \textbf{2,916} & \textbf{693} & \textbf{Real} & \textbf{Sports (Soccer)} & \textbf{Semantic Retrieval} \\
\bottomrule
\end{tabular}%
}
\caption{Comparison of video-to-video retrieval datasets. Our benchmark is the first to target semantic similarity in Soccer.}
\label{tab:datasets}
\end{table}

\begin{figure}
    \centering
    \begin{minipage}[t]{0.49\textwidth}
        \centering
        \includegraphics[width=\linewidth]{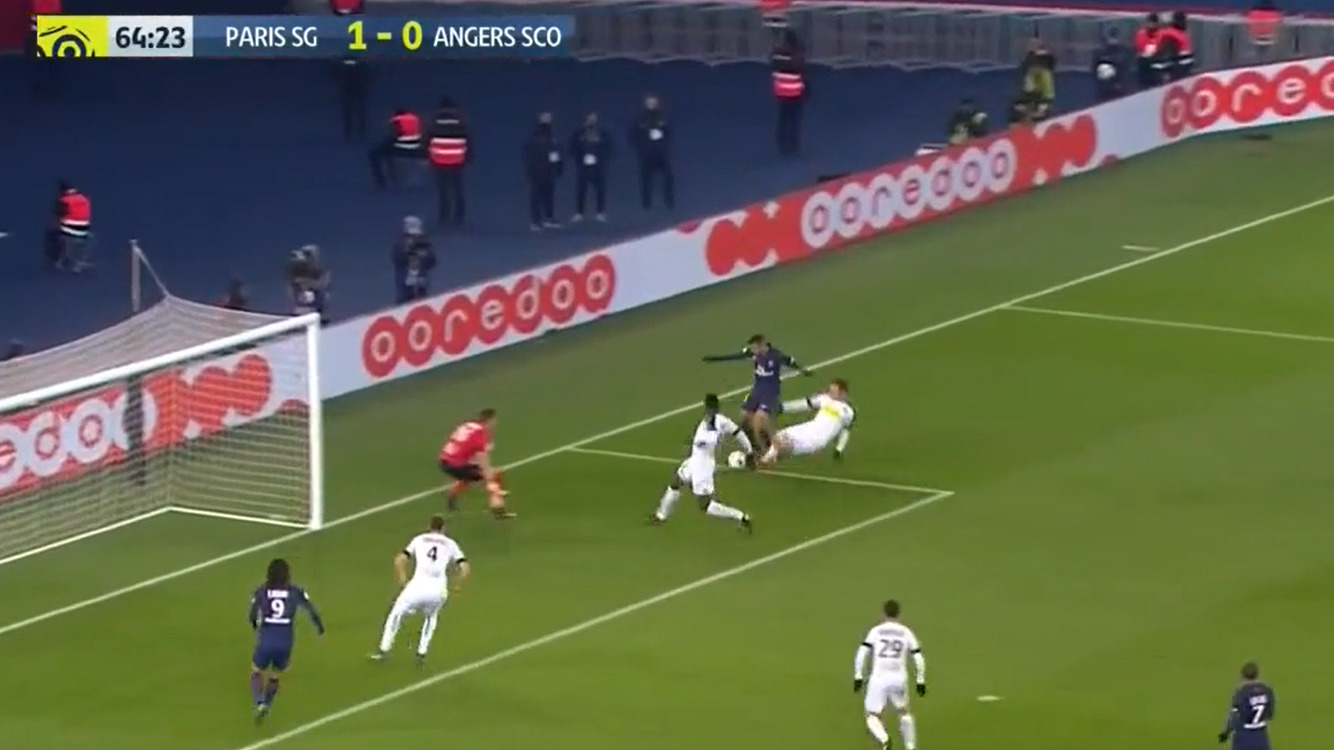}
        \subcaption{Tackle in the penalty area, stopping a goal-scoring opportunity. Awarded as a penalty.}
    \end{minipage}\hfill
    \begin{minipage}[t]{0.49\textwidth}
        \centering
        \includegraphics[width=\linewidth]{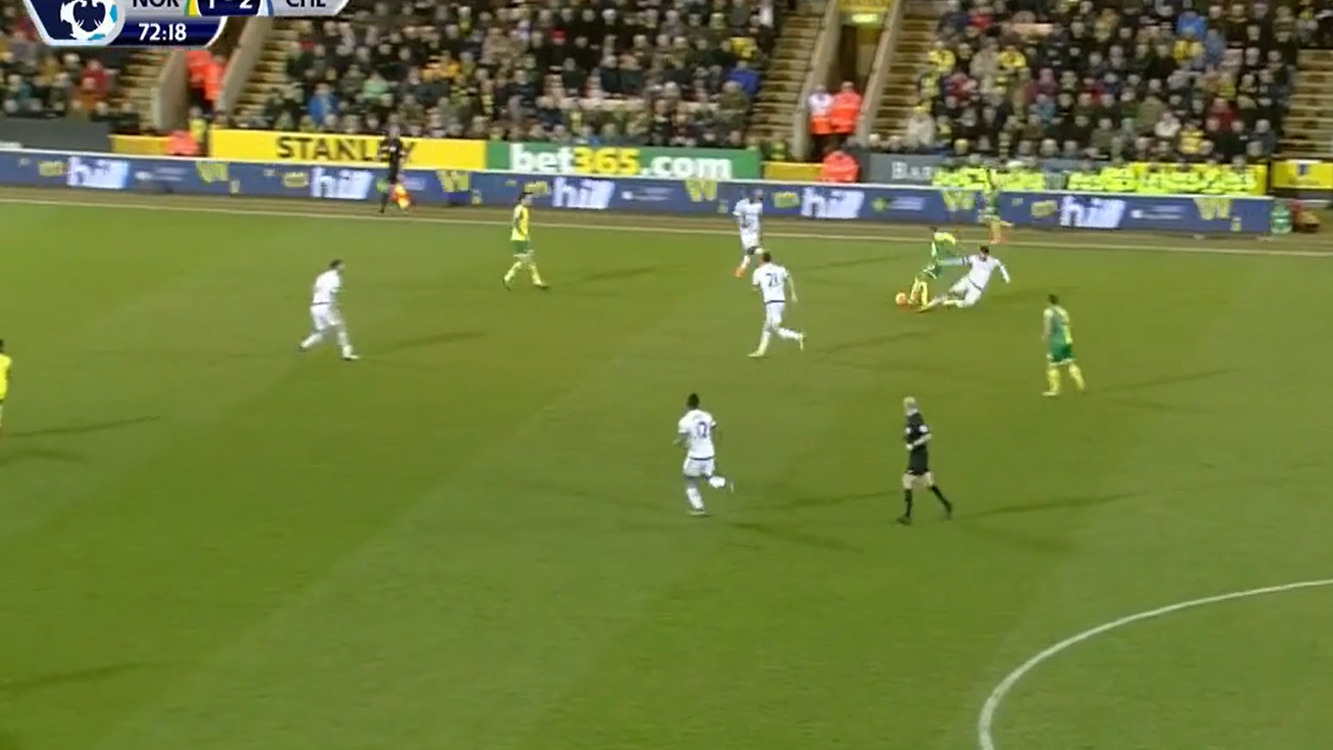}
        \subcaption{Tackle in midfield, disrupting an attack with no immediate goal-scoring opportunity. Awarded as a free kick.}
    \end{minipage}
    \caption{Two fouls with identical mechanics but vastly different sanctions depending on pitch location and context (penalty area vs. midfield).}
    \label{fig:similar_but_different}
\end{figure}

\section{The SoccerNet-FoulRet Benchmark}
\label{sec:Dataset_construction}
\label{sec:benchmark}
 
SoccerNet-FoulRet repurposes the SoccerNet-MVFoul~\cite{Held2023VARS} and SoccerNet-XFoul~\cite{Held2024XVARS} datasets into the first benchmark for \emph{semantic} foul retrieval. This section defines the task, describes the retrieval gallery, and details the two complementary evaluation targets: a human-verified set of precedent pairs, and a category-relevance evaluation.
 
\subsection{Task Definition}
The input is a query foul video $v_q$. The task is to rank a gallery of candidate clips so that relevant foul precedents appear first. We evaluate this retrieval task using two complementary notions of relevance. \textbf{Precedent relevance} is determined by Similar/Not Similar referee judgments of query--candidate pairs (\cref{tab:gold}). \textbf{Category relevance} is determined from the SoccerNet-MVFoul labels, at increasingly strict levels: Action, Action{+}Offence, and Action{+}Offence{+}Severity (\cref{tab:setB}). These two targets are evaluated separately throughout the paper. A legal slide tackle and a red-card lunge may look near-identical yet count as dissimilar, while two elbowing offences filmed from opposite ends of the pitch count as similar. This definition departs from every prior video-to-video retrieval benchmark (\cref{tab:datasets}) which ties relevance to visual or shared-event overlap.
 
\subsection{Retrieval Gallery}
\label{sec:construction}
The gallery is built exclusively from the training splits of SoccerNet-MVFoul
\cite{Held2023VARS} and SoccerNet-XFoul \cite{Held2024XVARS}, and comprises
$2{,}916$ actions. Alongside each action's video we retain two annotation signals.
First, its categorical attributes from MVFoul, including Action Class
(\eg, Tackle, Push), Offence, and Severity, which are used by the VARS+XVARS
baseline for coarse filtering and in our category-level evaluation. Second, its
XFoul question-answer (Q/A) pairs, which the VARS+XVARS baseline keeps as
separate semantic units rather than concatenating into one string, so that
similarity can be assessed dimension by dimension. The zero-shot video and
vision-language retrievers use neither signal because they take the raw clip as input
and produce their own embedding.

\subsection{Human-Verified Precedent Pairs}
The source datasets contain no similarity links between actions, so we build a verified \emph{stare decisis} precedent base over the validation and test splits of \cite{Held2023VARS} and \cite{Held2024XVARS}. We treat each validation/test action as a query against the training-split gallery, review its candidates as described below, and retain the $693$ queries ($293$ test\,/\,$400$ validation) for which at least one candidate is approved as a relevant precedent. These form the query set for our human-verified precedent evaluation. This subset is smaller than the full val+test pool (712 queries) because queries whose candidates are all rejected are dropped.
 
\begin{figure}
    \centering
    \includegraphics[width=1.0\linewidth]{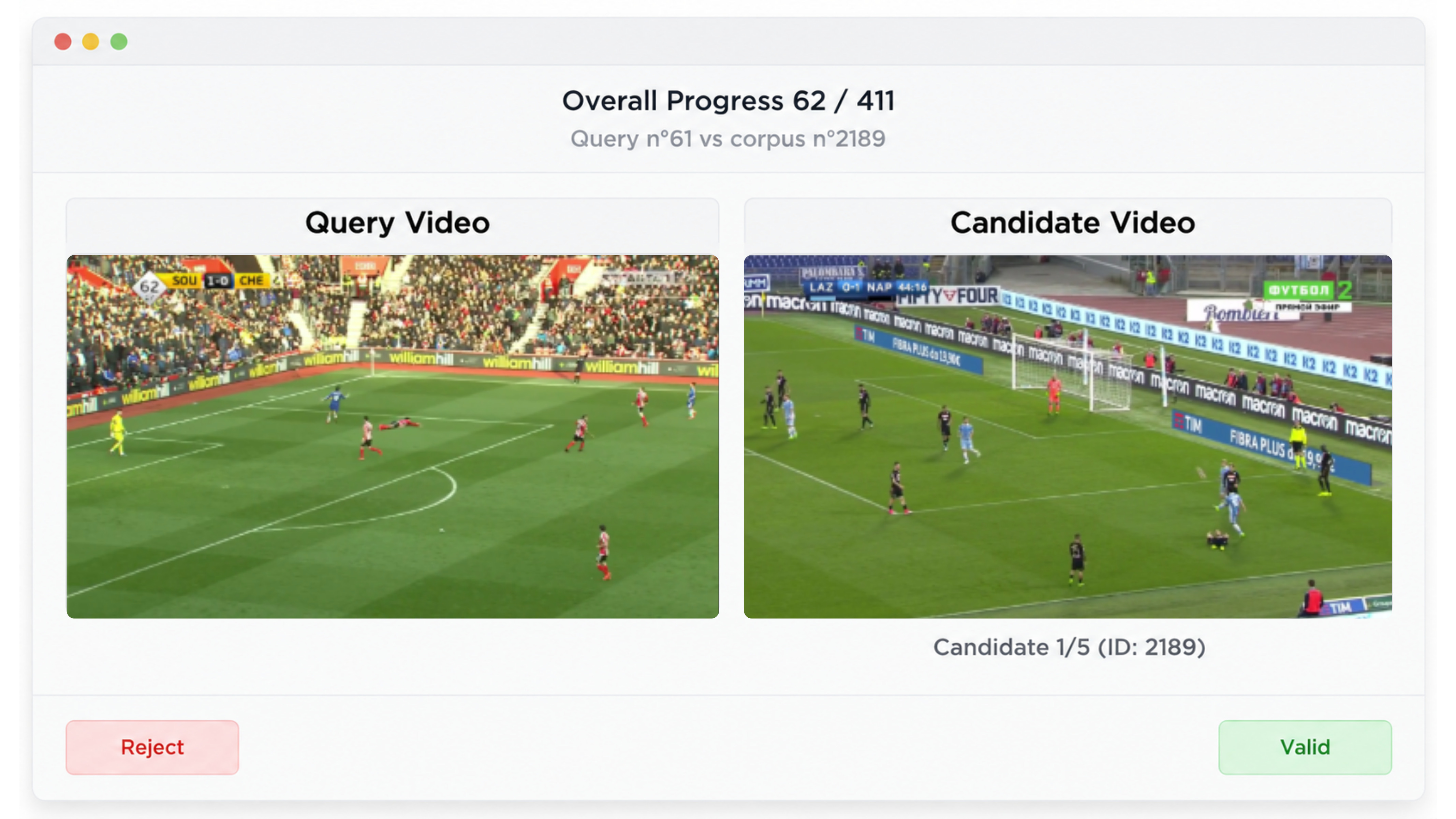}
    \caption{\textbf{Manual Validation Interface.} The annotator reviews a query action (left) alongside a candidate precedent (right), labeling each pair as \textit{Valid} or \textit{Rejected}.}
    \label{fig: GUI}
\end{figure}

\noindent\textbf{Oracle candidate generation.} For each query we display five candidates for human review. More details about the candidate-generation procedure is in \cref{sec:supp_candidate}. \Cref{fig: GUI} shows the graphical user interface annotators used to validate or reject each candidate clip as semantically similar to the query.

\noindent\textbf{Human validation.} Through a custom side-by-side interface, an annotator compares each query with its five candidates and labels every pair \emph{Valid} (a relevant precedent) or \emph{Rejected}. The full annotation criteria and protocol are provided in
\cref{sec:supp_annotation}. Because at most five candidates are reviewed per query, these judgments are a high-precision but \emph{incomplete} sample of all relevant precedents in the gallery, a property that dictates our choice of metrics (\cref{sec:Evaluation Metrics}). For each clip we additionally take its SoccerNet-MVFoul~\cite{Held2023VARS} attribute labels (Action type, Offence, and Severity), which we use as a complementary relevance signal in evaluation (\cref{sec:Evaluation Metrics}). The authors themselves served as annotators. Given their refereeing and soccer-video-understanding expertise, we found this sufficient to reliably determine whether each query–candidate pair is a valid precedent. Additional agreement analysis is provided in \cref{sec:supp_agreement}.

The complete SoccerNet-FoulRet benchmark comprises $693$ query fouls, each paired with
human-verified precedents (about four on average), evaluated against a gallery of $2{,}916$ actions. Additional statistics on the benchmark's relevance structure are provided in \cref{sec:supp_relevance}. \Cref{tab:stats} summarises the benchmark.

\begin{table}[t]\centering\small\setlength{\tabcolsep}{4.5pt}
\begin{tabular}{lrrr@{\quad}|@{\quad}lrrr}
\toprule
\emph{Action class} & Gal. & Qry. & Prec. & \emph{Offence / Severity} & Gal. & Qry. & Prec. \\
\midrule
Standing tackling & $1{,}264$ & $291$ & $1{,}183$ & Offence     & $2{,}495$ & $613$ & $2{,}490$ \\
Tackling          & $448$     & $124$ & $532$     & No offence  & $324$     & $53$  & $162$ \\
Challenge         & $383$     & $89$  & $302$     & Between     & $96$      & $27$  & $99$ \\
Holding           & $361$     & $93$  & $373$     & \multicolumn{4}{c}{} \\
\cmidrule(lr){5-8}
Elbowing          & $178$     & $39$  & $155$     & Sev.\ 1     & $1{,}402$ & $355$ & $1{,}399$ \\
High leg          & $103$     & $24$  & $104$     & Sev.\ 2     & $403$     & $91$  & $367$ \\
Pushing           & $88$      & $20$  & $65$      & Sev.\ 3     & $687$     & $169$ & $728$ \\
Don't know        & $52$      & $5$   & $20$      & Sev.\ 4     & $44$      & $11$  & $40$ \\
Dive              & $28$      & $5$   & $13$      & Sev.\ 5     & $27$      & $8$   & $35$ \\
\bottomrule
\end{tabular}
\caption{\textbf{SoccerNet-FoulRet composition.} Gallery clips (Gal.), query fouls (Qry.), and
approved precedent pairs (Prec.) across the three attribute types. All are long-tailed:
\emph{standing tackling} is ${\sim}44\%$ of the gallery, $86\%$ of clips are an \emph{offence},
and severities $4$--$5$ are rare (${<}3\%$). Counts are over each attribute's labelled subset
(full gallery $2{,}916$).}
\label{tab:stats}
\end{table}

\section{Experiments}
\label{sec:experiments}

\subsection{Evaluation Protocol}
\label{sec:Evaluation Metrics}
We use rank-based metrics for the sparse human-verified precedent pool and mAP for the dense category labels.

\noindent\textbf{Human-verified precedent metrics.} Let $A$ be the approved set for a query and $R_K$ the top-$K$ retrieved clips. We report \textbf{HitRate@$K$}, the fraction of queries with at least one verified precedent in $R_K$; \textbf{Recall@$K$} $=|R_K \cap A|/|A|$, the fraction of approved clips retrieved; \textbf{nDCG@$K$}, which measures the discounted ranking of approved clips using binary relevance; and \textbf{MRR@10}, the mean reciprocal rank of the first approved clip within the top 10. We evaluate HitRate and Recall at $K\in\{1,5,10\}$ and nDCG at $K\in\{5,10\}$. Because only a small subset of gallery clips is judged for each query, the precedent pool is incomplete, which is a known challenge in retrieval evaluation~\cite{Buckley2004Incomplete}. As a result, retrieved clips outside the verified set may still be valid precedents, but they receive no credit under these metrics.
 
\noindent\textbf{Category-relevance metric.} For category relevance (\cref{sec:benchmark}) we report mean Average Precision, \textbf{mAP@$K$}, at each of the three strictness levels (Action, Action{+}Offence, Action{+}Offence{+}Severity).
 
\noindent\textbf{Query sets.} The human-verified precedent metrics are computed over
the $693$ queries with at least one verified precedent. The category-relevance metrics
use all queries carrying the labels required at each strictness level.

\subsection{Implementation Details}
\label{sec:implementation}
We evaluate two retrieval families: a text-mediated VARS+XVARS pipeline and
video/vision--language embedding models. Within the latter, we evaluate both
zero-shot models and a task-specific adaptation of Qwen3-VL-Embedding-2B.
The embedding-based models operate directly on the video and rank gallery clips
by cosine similarity, while the VARS+XVARS pipeline uses predicted foul
attributes for filtering and generated question--answer pairs for semantic ranking.
We detail these retrieval settings below.

\noindent\textbf{Visual retrieval via generic vision-language encoders.} For the zero-shot baselines, we use existing off-the-shelf large vision-language encoders without fine-tuning on soccer data, though these models have most likely been exposed to sports footage during pre-training.
InternVideo2~\cite{wang2024internvideo2} (1B and 6B) is a video-native foundation model
trained with masked video modeling and cross-modal contrastive learning. It embeds a clip
directly and takes no text prompt. VLM2Vec-V2 \cite{Jiang2025VLM2Vec} and the
Qwen3-VL-Embedding series~\cite{Li2026Qwen3VLEmbedding-arxiv} (2B and 8B) turn Qwen2-VL and Qwen3-VL into unified multimodal encoders, and LamRA~\cite{liu2025lamra} re-purposes a generative multimodal model for retrieval through lightweight adapters. We additionally adapt Qwen3-VL-Embedding-2B to SoccerNet-MVFoul using LoRA
and symmetric hierarchical multi-positive InfoNCE objectives over Action,
Action+Offence, and Action+Offence+Severity. Fine-tuning uses only the SoccerNet-MVFoul training split. The full objective and training details are provided in \cref{sec:supp_finetune}.

\noindent\textbf{Textual retrieval via automated referee system-generated text.}
This text-mediated baseline chains the SoccerNet models. VARS~\cite{Held2023VARS} predicts the Action and Offence attributes for each query and gallery foul instance. We restrict the search space to gallery instances whose two predicted attributes exactly match those of the query. XVARS~\cite{Held2024XVARS} then generates a textual profile using four of the original refereeing questions from \cite{Held2024XVARS}, shown below:

\begin{promptbox}[XVARS Prompt]
\small
You are an artificial intelligence assistant for visual football referee questions.
Give short and helpful answers to the user's questions:

\medskip
\noindent
``What card would you give? Why?'';\\
``Did the defender touch the ball?'';\\
``Did the defender intend to play the ball?'';\\
``Where was the point of contact?''
\end{promptbox}

For each of the four question fields, we encode the generated answer of the
query with BGE~\cite{Xiao2024CPack}
(\texttt{bge-large-en-v1.5}) and compare it with the corresponding
precomputed BGE answer embedding of each candidate. We average the four
per-answer cosine similarities,
\begin{equation}
\label{eq:avgsim}
\mathrm{Score}(Q, C) = \frac{1}{N} \sum_{i=1}^{N} \frac{\vec{v}_{q_i} \cdot \vec{v}_{c_i}}{\lVert \vec{v}_{q_i}\rVert\,\lVert \vec{v}_{c_i}\rVert},
\end{equation}
where $\vec{v}_{q_i}$ and $\vec{v}_{c_i}$ are the BGE embeddings of the
generated answers for the $i$-th question field of query $Q$ and candidate
$C$, respectively, and $N=4$. Averaging per answer, rather than embedding
one concatenated description, forces agreement across multiple semantic
dimensions instead of rewarding generic textual overlap.

\noindent\textbf{Instruction protocol.} 
The instruction-following embedders (VLM2Vec-V2, Qwen3-VL, LamRA) receive an identical query-side instruction, ``Retrieve videos similar to the query video.'', each in its own native template. Following how these models are designed, it is applied to the query only, and gallery clips carry no task instruction (VLM2Vec-V2 and Qwen3-VL use their default prompt, LamRA uses its native ``summarise in one word'' template). InternVideo2 is video-native and uses no instruction. We ablate instruction specificity in \cref{sec:ablation}.

\subsection{Results}

\noindent\textbf{Gold precedents.} We use this benchmark to quantify how well
off-the-shelf retrievers perform on foul precedent retrieval and whether task-specific adaptation improves performance. We also evaluate the task-specific Qwen3-VL-Embedding-2B baseline described in \cref{sec:implementation}. We therefore read every method against its random floor rather than in absolute terms.

\begin{table}[t]
\centering\small
\setlength{\tabcolsep}{2pt}
\renewcommand{\arraystretch}{1.15}
\resizebox{\columnwidth}{!}{%
\begin{tabular}{l c ccc ccc cc c}
\toprule
& & \multicolumn{3}{c}{\textbf{HitRate@$K$}}
& \multicolumn{3}{c}{\textbf{Recall@$K$}}
& \multicolumn{2}{c}{\textbf{nDCG@$K$}} & \\
\cmidrule(lr){3-5}\cmidrule(lr){6-8}\cmidrule(lr){9-10}
Method & Params & @1 & @5 & @10 & @1 & @5 & @10
& @5 & @10 & MRR\@10 \\
\midrule
Random & --- & 0.14 & 0.68 & 1.35
& 0.03 & 0.17 & 0.34
& 0.16 & 0.24 & 0.40 \\

\midrule
\multicolumn{11}{c}{\textit{Text-mediated retrieval}} \\
\midrule
VARS+XVARS & ---
& \shortstack{0.14\\[-1pt]{\tiny[0.00,0.43]}}
& \shortstack{2.16\\[-1pt]{\tiny[1.15,3.32]}}
& \shortstack{3.17\\[-1pt]{\tiny[1.88,4.47]}}
& \shortstack{0.03\\[-1pt]{\tiny[0.00,0.09]}}
& \shortstack{0.64\\[-1pt]{\tiny[0.32,1.00]}}
& \shortstack{0.94\\[-1pt]{\tiny[0.56,1.38]}}
& \shortstack{0.51\\[-1pt]{\tiny[0.26,0.81]}}
& \shortstack{0.65\\[-1pt]{\tiny[0.38,0.96]}}
& \shortstack{0.99\\[-1pt]{\tiny[0.54,1.51]}} \\

\midrule
\multicolumn{11}{c}{\textit{Video-native retrieval}} \\
\midrule
InternVideo2-1B & 1B
& \shortstack{0.72\\[-1pt]{\tiny[0.14,1.44]}}
& \shortstack{\textbf{2.60}\\[-1pt]{\tiny[1.44,3.90]}}
& \shortstack{\textbf{4.62}\\[-1pt]{\tiny[3.03,6.20]}}
& \shortstack{0.15\\[-1pt]{\tiny[0.03,0.30]}}
& \shortstack{\textbf{0.66}\\[-1pt]{\tiny[0.36,1.00]}}
& \shortstack{\textbf{1.13}\\[-1pt]{\tiny[0.74,1.55]}}
& \shortstack{\textbf{0.62}\\[-1pt]{\tiny[0.34,0.94]}}
& \shortstack{\textbf{0.87}\\[-1pt]{\tiny[0.56,1.21]}}
& \shortstack{\textbf{1.65}\\[-1pt]{\tiny[0.96,2.44]}} \\

InternVideo2-6B & 6B
& \shortstack{0.29\\[-1pt]{\tiny[0.00,0.72]}}
& \shortstack{1.59\\[-1pt]{\tiny[0.72,2.60]}}
& \shortstack{3.46\\[-1pt]{\tiny[2.16,4.91]}}
& \shortstack{0.06\\[-1pt]{\tiny[0.00,0.17]}}
& \shortstack{0.43\\[-1pt]{\tiny[0.18,0.73]}}
& \shortstack{1.03\\[-1pt]{\tiny[0.60,1.55]}}
& \shortstack{0.35\\[-1pt]{\tiny[0.15,0.58]}}
& \shortstack{0.61\\[-1pt]{\tiny[0.36,0.89]}}
& \shortstack{0.88\\[-1pt]{\tiny[0.45,1.41]}} \\

\midrule
\multicolumn{11}{c}{\textit{Instruction-following multimodal retrieval}} \\
\midrule
Qwen3-VL-Emb-2B-FT & 2B
& \shortstack{0.58\\[-1pt]{\tiny[0.14,1.15]}}
& \shortstack{2.02\\[-1pt]{\tiny[1.01,3.17]}}
& \shortstack{3.75\\[-1pt]{\tiny[2.45,5.19]}}
& \shortstack{0.18\\[-1pt]{\tiny[0.03,0.38]}}
& \shortstack{0.57\\[-1pt]{\tiny[0.28,0.92]}}
& \shortstack{1.06\\[-1pt]{\tiny[0.63,1.57]}}
& \shortstack{0.55\\[-1pt]{\tiny[0.26,0.89]}}
& \shortstack{0.77\\[-1pt]{\tiny[0.45,1.13]}}
& \shortstack{1.24\\[-1pt]{\tiny[0.66,1.92]}} \\

Qwen3-VL-Emb-2B & 2B
& \shortstack{\textbf{1.01}\\[-1pt]{\tiny[0.29,1.73]}}
& \shortstack{1.73\\[-1pt]{\tiny[0.87,2.74]}}
& \shortstack{3.32\\[-1pt]{\tiny[2.02,4.76]}}
& \shortstack{\textbf{0.23}\\[-1pt]{\tiny[0.08,0.41]}}
& \shortstack{0.37\\[-1pt]{\tiny[0.17,0.60]}}
& \shortstack{0.95\\[-1pt]{\tiny[0.54,1.44]}}
& \shortstack{0.50\\[-1pt]{\tiny[0.23,0.82]}}
& \shortstack{0.76\\[-1pt]{\tiny[0.44,1.11]}}
& \shortstack{1.52\\[-1pt]{\tiny[0.78,2.37]}} \\

VLM2Vec-V2 & 2B
& \shortstack{0.72\\[-1pt]{\tiny[0.14,1.44]}}
& \shortstack{2.02\\[-1pt]{\tiny[1.01,3.17]}}
& \shortstack{3.90\\[-1pt]{\tiny[2.45,5.34]}}
& \shortstack{0.20\\[-1pt]{\tiny[0.04,0.39]}}
& \shortstack{0.59\\[-1pt]{\tiny[0.26,1.01]}}
& \shortstack{1.06\\[-1pt]{\tiny[0.64,1.55]}}
& \shortstack{0.58\\[-1pt]{\tiny[0.28,0.94]}}
& \shortstack{0.81\\[-1pt]{\tiny[0.48,1.19]}}
& \shortstack{1.42\\[-1pt]{\tiny[0.76,2.17]}} \\

LamRA & 7B
& \shortstack{0.72\\[-1pt]{\tiny[0.14,1.44]}}
& \shortstack{1.73\\[-1pt]{\tiny[0.87,2.74]}}
& \shortstack{3.17\\[-1pt]{\tiny[1.88,4.47]}}
& \shortstack{0.18\\[-1pt]{\tiny[0.04,0.35]}}
& \shortstack{0.52\\[-1pt]{\tiny[0.21,0.93]}}
& \shortstack{0.93\\[-1pt]{\tiny[0.52,1.42]}}
& \shortstack{0.49\\[-1pt]{\tiny[0.22,0.80]}}
& \shortstack{0.68\\[-1pt]{\tiny[0.38,1.01]}}
& \shortstack{1.20\\[-1pt]{\tiny[0.59,1.92]}} \\

Qwen3-VL-Emb-8B & 8B
& \shortstack{0.43\\[-1pt]{\tiny[0.00,1.01]}}
& \shortstack{1.73\\[-1pt]{\tiny[0.87,2.74]}}
& \shortstack{3.17\\[-1pt]{\tiny[1.88,4.47]}}
& \shortstack{0.11\\[-1pt]{\tiny[0.00,0.26]}}
& \shortstack{0.65\\[-1pt]{\tiny[0.25,1.15]}}
& \shortstack{1.00\\[-1pt]{\tiny[0.53,1.56]}}
& \shortstack{0.51\\[-1pt]{\tiny[0.23,0.84]}}
& \shortstack{0.69\\[-1pt]{\tiny[0.38,1.04]}}
& \shortstack{1.08\\[-1pt]{\tiny[0.55,1.72]}} \\
\bottomrule
\end{tabular}
}
\caption{Retrieval performance on the human-verified precedent pool,
pooled over the $693$ queries that have at least one
verified precedent. \textbf{HitRate@$K$} is the fraction of queries with at
least one verified precedent in the top $K$; \textbf{Recall@$K$} is the
fraction of verified precedents retrieved; \textbf{nDCG@$K$} measures the
discounted ranking of the available verified precedents using binary
relevance; and \textbf{MRR@10} is the mean reciprocal rank of the first
verified precedent within the top 10. All values are percentages.
Because the relevance pool is incomplete, unjudged retrieved clips may also
be valid precedents and receive no relevance gain under these metrics.
Qwen3-VL-Emb-2B-FT is fine-tuned exclusively on SoccerNet-MVFoul's
train set. All other visual embedders are evaluated zero-shot.
Higher is better. Brackets report 95\% percentile-bootstrap confidence intervals obtained
from $20{,}000$ resamples. \textbf{Bold} denotes the best result per column.}
\label{tab:gold}
\end{table}

\Cref{tab:gold} shows that InternVideo2-1B, a zero-shot model applied with no domain-specific
training, is the strongest retriever overall. It leads at the shallow cutoffs a user actually
inspects (HR@5 $2.60\%$, HR@10 $4.62\%$), on Recall (R@5 $0.66\%$, R@10 $1.13\%$), on nDCG (nDCG@5 $0.62\%$, nDCG@10 $0.87\%$), and on MRR@10 ($1.65$). 

The engineered VARS{+}XVARS pipeline leads on no metric. Most tellingly, at rank~1 the pipeline sits exactly on the random floor (HR@1 $0.14\%$). This indicates that VARS' coarse category filter can narrow the pool without reliably placing a human-verified precedent at the top, underscoring that category agreement alone does not imply precedent relevance. The pipeline also returns no candidates for $17\%$ of Test queries, which are scored as misses. 

In contrast, the embedders place a verified precedent at rank~1 more often than
VARS+XVARS, led by Qwen3-VL-2B (HR@1 $1.01\%$ and R@1 $0.23\%$). Recall is consistently much lower than HitRate. For InternVideo2-1B, for example, HR@10 is $4.62\%$ while R@10 is only $1.13\%$, indicating that even successful top-10 retrievals recover only a small fraction of the approximately four verified precedents available per query.

Among these zero-shot embedders, two further patterns emerge. First, scaling does not improve retrieval within either model family (InternVideo2 and Qwen3-VL-Emb). InternVideo2-6B trails InternVideo2-1B, while Qwen3-VL-Emb-8B does not outperform Qwen3-VL-Emb-2B. InternVideo2-1B remains the strongest overall zero-shot retriever, whereas InternVideo2-6B records the lowest MRR@10 among the zero-shot embedders ($0.88\%$). This suggests that increasing model size
alone does not yield better transfer to foul retrieval in our setting. Second, at HR@10 every method exceeds the random baseline, but only by $2.3$--$3.4\times$, showing how far the task remains from solved.

Task-specific fine-tuning produces mixed results on the human-verified
precedent evaluation. Qwen3-VL-Emb-2B-FT improves HR, Recall, and nDCG at
deeper cutoffs, but reduces HR@1 and MRR@10 relative to its zero-shot
initialization. Thus, category supervision improves deeper retrieval without
improving the earliest result, suggesting that action, offence, and severity
do not fully capture precedent relevance.

\noindent\textbf{Category relevance.} The human-verified precedent results above
should be interpreted in light of the sparse relevance pool. Only about five of the
$2{,}916$ gallery clips were reviewed per query, so unjudged retrieved clips may still
be valid precedents but receive no credit under these metrics. To score relevance independently of this pool, we turn to the official SoccerNet-MVFoul \cite{Held2023VARS} attribute labels, which cover the entire gallery and let us measure ranking quality at three nested strictness levels (Action, Action{+}Offence, Action{+}Offence{+}Severity), shown in \cref{tab:setB}.

\begin{table}[t]
\centering\small
\setlength{\tabcolsep}{2pt}
\renewcommand{\arraystretch}{1.15}
\resizebox{\columnwidth}{!}{%
\begin{tabular}{l c *{9}{c}}
\toprule
& & \multicolumn{3}{c}{Action}
& \multicolumn{3}{c}{{+}Offence}
& \multicolumn{3}{c}{{+}Severity$^{*}$} \\
\cmidrule(lr){3-5}\cmidrule(lr){6-8}\cmidrule(lr){9-11}
Model & Params & @1 & @5 & @10 & @1 & @5 & @10 & @1 & @5 & @10 \\
\midrule
Random & ---
& 24.9 & 16.1 & 13.5
& 20.8 & 13.3 & 11.0
& 11.0 & 6.2 & 4.8 \\

\midrule
\multicolumn{11}{c}{\textit{Text-mediated retrieval}} \\
\midrule
VARS+XVARS$^{\dagger}$ & ---
& \shortstack{38.5\\[-1pt]{\tiny[35.0,42.2]}}
& \shortstack{29.4\\[-1pt]{\tiny[27.1,31.7]}}
& \shortstack{29.0\\[-1pt]{\tiny[26.8,31.2]}}
& \shortstack{32.7\\[-1pt]{\tiny[29.2,36.2]}}
& \shortstack{24.5\\[-1pt]{\tiny[22.4,26.7]}}
& \shortstack{24.0\\[-1pt]{\tiny[21.9,26.2]}}
& \shortstack{20.6\\[-1pt]{\tiny[17.5,23.8]}}
& \shortstack{13.5\\[-1pt]{\tiny[11.8,15.2]}}
& \shortstack{10.7\\[-1pt]{\tiny[9.5,11.9]}} \\

\midrule
\multicolumn{11}{c}{\textit{Video-native retrieval}} \\
\midrule
InternVideo2-1B & 1B
& \shortstack{37.5\\[-1pt]{\tiny[34.0,41.2]}}
& \shortstack{26.3\\[-1pt]{\tiny[24.3,28.4]}}
& \shortstack{23.1\\[-1pt]{\tiny[21.5,24.8]}}
& \shortstack{29.9\\[-1pt]{\tiny[26.7,33.3]}}
& \shortstack{20.9\\[-1pt]{\tiny[19.0,22.8]}}
& \shortstack{18.3\\[-1pt]{\tiny[16.8,20.0]}}
& \shortstack{17.1\\[-1pt]{\tiny[14.3,20.0]}}
& \shortstack{9.5\\[-1pt]{\tiny[8.4,10.8]}}
& \shortstack{7.4\\[-1pt]{\tiny[6.5,8.2]}} \\

InternVideo2-6B & 6B
& \shortstack{37.8\\[-1pt]{\tiny[34.3,41.3]}}
& \shortstack{25.9\\[-1pt]{\tiny[23.9,27.9]}}
& \shortstack{22.2\\[-1pt]{\tiny[20.5,23.9]}}
& \shortstack{30.7\\[-1pt]{\tiny[27.4,34.3]}}
& \shortstack{20.4\\[-1pt]{\tiny[18.5,22.3]}}
& \shortstack{17.5\\[-1pt]{\tiny[15.9,19.1]}}
& \shortstack{\textbf{17.7}\\[-1pt]{\tiny[14.8,20.8]}}
& \shortstack{8.9\\[-1pt]{\tiny[7.8,10.0]}}
& \shortstack{7.0\\[-1pt]{\tiny[6.3,7.9]}} \\

\midrule
\multicolumn{11}{c}{\textit{Instruction-following multimodal retrieval}} \\
\midrule
Qwen3-VL-Emb-2B-FT & 2B
& \shortstack{\textbf{40.3}\\[-1pt]{\tiny[36.7,44.0]}}
& \shortstack{\textbf{31.9}\\[-1pt]{\tiny[29.5,34.3]}}
& \shortstack{\textbf{29.0}\\[-1pt]{\tiny[26.9,31.1]}}
& \shortstack{\textbf{33.7}\\[-1pt]{\tiny[30.2,37.1]}}
& \shortstack{\textbf{27.0}\\[-1pt]{\tiny[24.8,29.3]}}
& \shortstack{\textbf{24.0}\\[-1pt]{\tiny[22.1,26.0]}}
& \shortstack{\textbf{17.7}\\[-1pt]{\tiny[14.8,20.6]}}
& \shortstack{\textbf{10.9}\\[-1pt]{\tiny[9.7,12.2]}}
& \shortstack{\textbf{8.7}\\[-1pt]{\tiny[7.8,9.7]}} \\

Qwen3-VL-Emb-2B & 2B
& \shortstack{37.9\\[-1pt]{\tiny[34.4,41.5]}}
& \shortstack{27.1\\[-1pt]{\tiny[25.1,29.3]}}
& \shortstack{23.0\\[-1pt]{\tiny[21.4,24.7]}}
& \shortstack{32.4\\[-1pt]{\tiny[29.1,36.0]}}
& \shortstack{22.6\\[-1pt]{\tiny[20.6,24.6]}}
& \shortstack{18.9\\[-1pt]{\tiny[17.3,20.5]}}
& \shortstack{16.5\\[-1pt]{\tiny[13.7,19.4]}}
& \shortstack{9.2\\[-1pt]{\tiny[8.1,10.4]}}
& \shortstack{7.1\\[-1pt]{\tiny[6.3,7.9]}} \\

VLM2Vec-V2 & 2B
& \shortstack{33.3\\[-1pt]{\tiny[29.8,36.7]}}
& \shortstack{22.2\\[-1pt]{\tiny[20.4,24.0]}}
& \shortstack{18.8\\[-1pt]{\tiny[17.4,20.3]}}
& \shortstack{28.1\\[-1pt]{\tiny[24.8,31.5]}}
& \shortstack{18.3\\[-1pt]{\tiny[16.6,20.0]}}
& \shortstack{15.5\\[-1pt]{\tiny[14.1,16.8]}}
& \shortstack{13.5\\[-1pt]{\tiny[10.9,16.2]}}
& \shortstack{8.0\\[-1pt]{\tiny[7.0,9.1]}}
& \shortstack{6.3\\[-1pt]{\tiny[5.5,7.1]}} \\

LamRA & 7B
& \shortstack{34.7\\[-1pt]{\tiny[31.3,38.2]}}
& \shortstack{22.2\\[-1pt]{\tiny[20.4,24.1]}}
& \shortstack{18.7\\[-1pt]{\tiny[17.3,20.2]}}
& \shortstack{30.2\\[-1pt]{\tiny[26.8,33.4]}}
& \shortstack{19.3\\[-1pt]{\tiny[17.5,21.1]}}
& \shortstack{15.9\\[-1pt]{\tiny[14.5,17.3]}}
& \shortstack{13.1\\[-1pt]{\tiny[10.6,15.7]}}
& \shortstack{7.2\\[-1pt]{\tiny[6.1,8.3]}}
& \shortstack{5.6\\[-1pt]{\tiny[4.8,6.4]}} \\

Qwen3-VL-Emb-8B & 8B
& \shortstack{35.3\\[-1pt]{\tiny[31.7,38.8]}}
& \shortstack{23.8\\[-1pt]{\tiny[22.0,25.6]}}
& \shortstack{20.5\\[-1pt]{\tiny[19.0,22.0]}}
& \shortstack{29.5\\[-1pt]{\tiny[26.1,32.9]}}
& \shortstack{19.1\\[-1pt]{\tiny[17.5,20.9]}}
& \shortstack{16.3\\[-1pt]{\tiny[15.0,17.7]}}
& \shortstack{15.8\\[-1pt]{\tiny[13.1,18.8]}}
& \shortstack{8.3\\[-1pt]{\tiny[7.3,9.4]}}
& \shortstack{6.4\\[-1pt]{\tiny[5.7,7.2]}} \\
\bottomrule
\end{tabular}
}
\caption{Category relevance, scored with the official SoccerNet-MVFoul attribute
labels. A retrieved clip is relevant if it matches the query's foul at one of three increasingly
strict levels: Action, Action{+}Offence, or Action{+}Offence{+}Severity verdict. For each level, both the retrieval gallery and evaluated
queries are restricted to clips carrying the required labels. Action and {+}Offence use
$2{,}905$ gallery clips and $709$ queries; {+}Severity uses $2{,}561$ gallery clips and
$650$ queries. Values are mean Average Precision (\textbf{mAP@$K$}, \%). Qwen3-VL-Emb-2B-FT is fine-tuned exclusively on SoccerNet-MVFoul's train set, whereas the other visual embedders are evaluated zero-shot. VARS+XVARS$^{\dagger}$ uses VARS-predicted categories for candidate filtering and is therefore excluded from the best-model comparison. $^{*}$Severity is annotated for only ${\sim}88\%$ of the gallery, so this tier contains fewer clips and queries. Higher is better. Brackets report 95\% percentile-bootstrap confidence intervals obtained from $20{,}000$ resamples. \textbf{Bold} denotes the best result per column.}
\label{tab:setB}
\end{table}

Among the zero-shot embedders, InternVideo2-1B and Qwen3-VL-2B are the strongest
overall on the category evaluation. Qwen3-VL-2B leads most Action and
Action{+}Offence metrics, while InternVideo2-1B is among the strongest zero-shot
models at the strictest Action{+}Offence{+}Severity level. InternVideo2-1B also
performs best on the human-verified precedent evaluation (\cref{tab:gold}), so the
strongest zero-shot models are broadly consistent across both evaluations.
Qwen3-VL-8B trails its 2B counterpart across all nine category metrics, suggesting
that greater model scale does not improve zero-shot transfer in this setting.

Performance decreases as the category criterion becomes stricter. For
Intern-Video2-1B, for example, mAP@10 drops from $23.1\%$ on Action to $18.3\%$
on {+}Offence and $7.4\%$ on {+}Severity. Qwen3-VL-Emb-2B-FT improves on its
zero-shot counterpart across all nine category metrics, showing that these labels
provide a useful supervision signal. VARS+XVARS also performs particularly well
on this evaluation, achieving the highest {+}Severity scores and tying
Qwen3-VL-Emb-2B-FT on Action mAP@10 ($29.0$) and {+}Offence mAP@10 ($24.0$).
This advantage is expected because the pipeline explicitly filters candidates using
VARS-predicted categories, which closely align with what these metrics reward.

However, these category-level gains do not translate directly to precedent retrieval.
VARS+XVARS remains weak on the human-verified precedent evaluation
(\cref{tab:gold}), while the task-specific baseline transfers only modestly despite
its substantial category improvements. Together, these results show that action,
offence, and severity capture useful structure but do not fully determine human
judgments of precedent relevance.

\subsection{Ablation}
\label{sec:ablation}

The instruction-following embedders take a query-side instruction, so we test whether
increasing task specificity affects retrieval. The prompts progressively add task-specific
information, from the retrieval objective, to soccer-foul context, to explicit category
labels, allowing us to test whether these cues affect zero-shot embeddings. We compare three levels of specificity, each applied in the model's native template: \textbf{Generic} (``Retrieve
videos similar to the query video.''), \textbf{Domain} (``Retrieve soccer foul videos similar
to the query.''), and \textbf{Detailed} (``Retrieve soccer foul videos of the same action
type, offence, and severity as the query.''). InternVideo2 is video-native and takes no
instruction, so it is excluded.

\begin{table}[t]\centering\small\setlength{\tabcolsep}{4pt}
\begin{tabular}{l l c ccc}
\toprule
& & \textbf{Gold Precedents} & \multicolumn{3}{c}{\textbf{Action}} \\
\cmidrule(lr){3-3}\cmidrule(lr){4-6}
Model & Instruction\ & MRR@10 & mAP@1 & mAP@5 & mAP@10 \\
\midrule
Qwen3-VL-Emb-2B & Generic  & \textbf{1.52} & 37.7 & \textbf{27.0} & \textbf{22.9} \\
                & Domain   & 1.15 & \textbf{39.1} & 26.6 & 22.1 \\
                & Detailed & 0.82 & 36.1 & 25.8 & 21.7 \\
\midrule
Qwen3-VL-Emb-8B & Generic  & \textbf{1.08} & \textbf{35.3} & \textbf{23.7} & \textbf{20.5} \\
                & Domain   & 0.66 & 30.3 & 19.7 & 16.6 \\
                & Detailed & 0.93 & 33.0 & 22.3 & 18.8 \\
\midrule
VLM2Vec-V2      & Generic  & \textbf{1.42} & \textbf{33.3} & \textbf{22.0} & \textbf{18.7} \\
                & Domain   & 1.33 & 29.6 & 19.8 & 16.9 \\
                & Detailed & 1.29 & 31.2 & 20.4 & 17.1 \\
\midrule
LamRA           & Generic  & \textbf{1.20} & \textbf{34.4} & \textbf{22.0} & \textbf{18.6} \\
                & Domain   & 0.32 & 25.4 & 17.9 & 15.1 \\
                & Detailed & 0.36 & 24.3 & 17.9 & 15.4 \\
\bottomrule
\end{tabular}
\caption{\textbf{Instruction-sensitivity ablation.} Query-side instruction specificity for the
four instruction-following embedders, from \textbf{Generic} to \textbf{Detailed}. We report
referee-gold MRR@10 and category mAP at the Action level (\%); higher is better. \textbf{Bold} =
best instruction per model and column. InternVideo2 is video-native and takes no instruction.}
\label{tab:ablation}
\end{table}

The Generic instruction is best or tied-best almost everywhere
(\cref{tab:ablation}), while adding soccer-specific context or explicit category
attributes generally does not improve performance. Notably, the Detailed instruction
names the evaluated category attributes yet never outperforms Generic on category mAP.
Within this limited progression of increasingly task-specific prompts, we therefore use
the Generic instruction for all main results. These results suggest that explicitly emphasizing the annotated foul categories in the instruction
does not improve the zero-shot embedding space, indicating that useful retrieval
signals are not captured by these attributes alone.

\subsection{Qualitative Analysis}
\label{sec:qualitative}

The two examples below illustrate a successful and a failed retrieval. Together, they
highlight both what the task requires and why it remains challenging.

\begin{figure}
    \centering
    \begin{minipage}{0.49\textwidth}
        \centering
        \includegraphics[width=\linewidth]{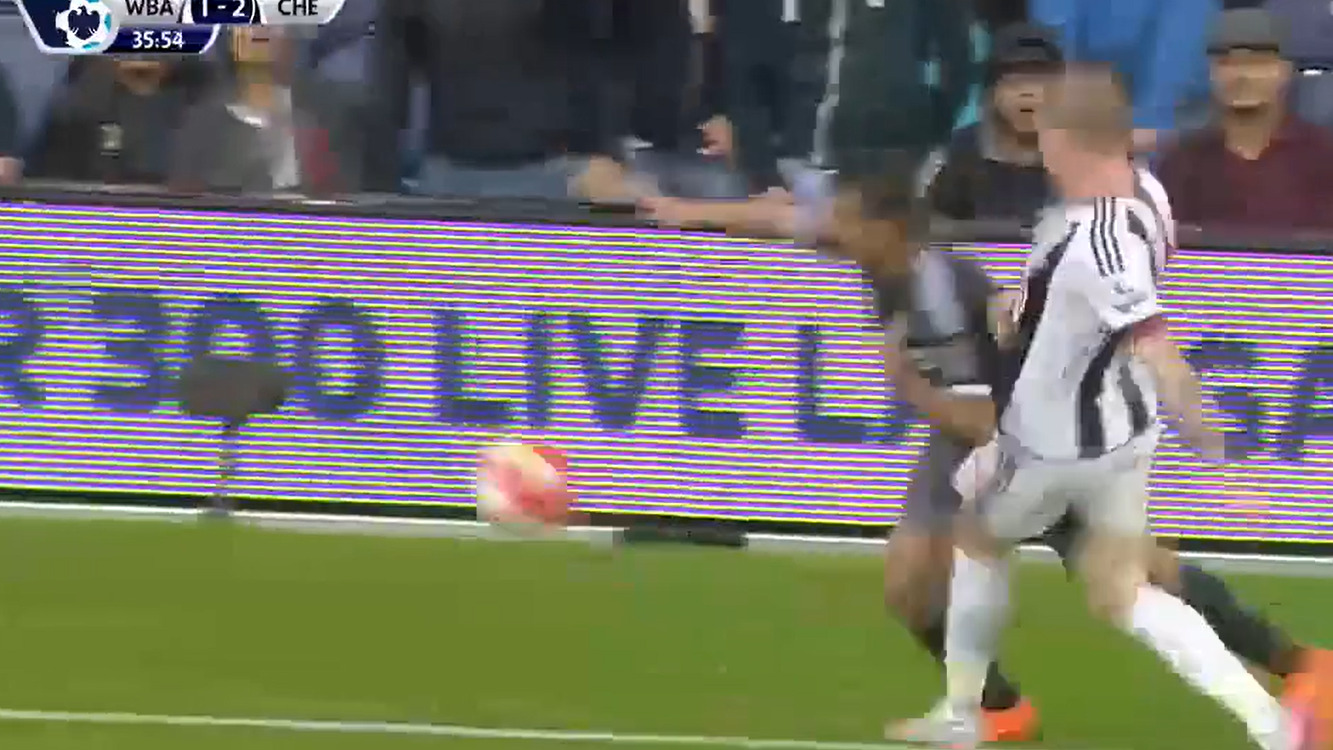}
        \subcaption{Query Video}
    \end{minipage}\hfill
    \begin{minipage}{0.49\textwidth}
        \centering
        \includegraphics[width=\linewidth]{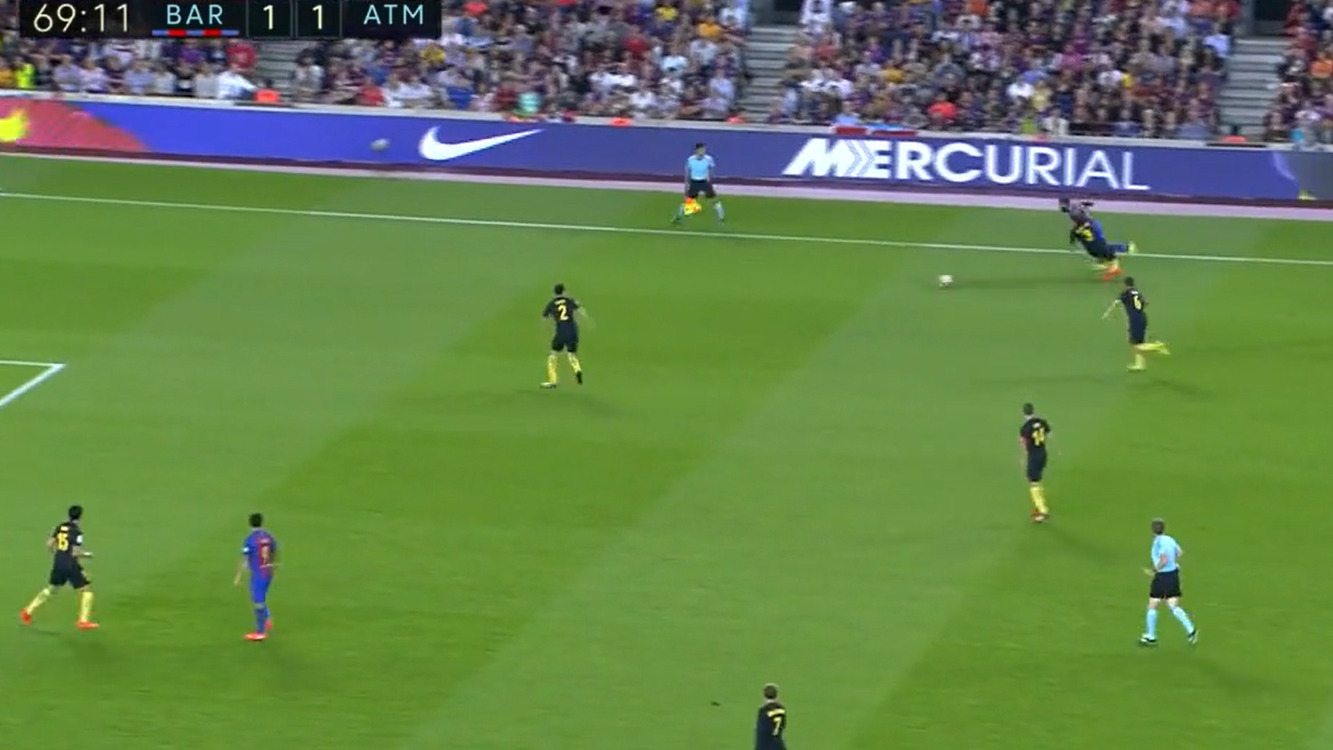}
        \subcaption{Retrieved Precedent}
    \end{minipage}
    \caption{A successful retrieval: a human-verified precedent is retrieved despite differences in camera zoom.}
    \label{fig:qualitative_success}
\end{figure}

\Cref{fig:qualitative_success} shows a successful retrieval. The query and retrieved
clip form a human-verified precedent pair despite substantial differences in camera angle,
lighting, and team kits. This illustrates that successful precedent retrieval requires
looking beyond visual appearance and that semantically relevant fouls can be matched
across visually different clips.

\begin{figure}
    \centering
    \begin{minipage}{0.49\textwidth}
        \centering
        \includegraphics[width=\linewidth]{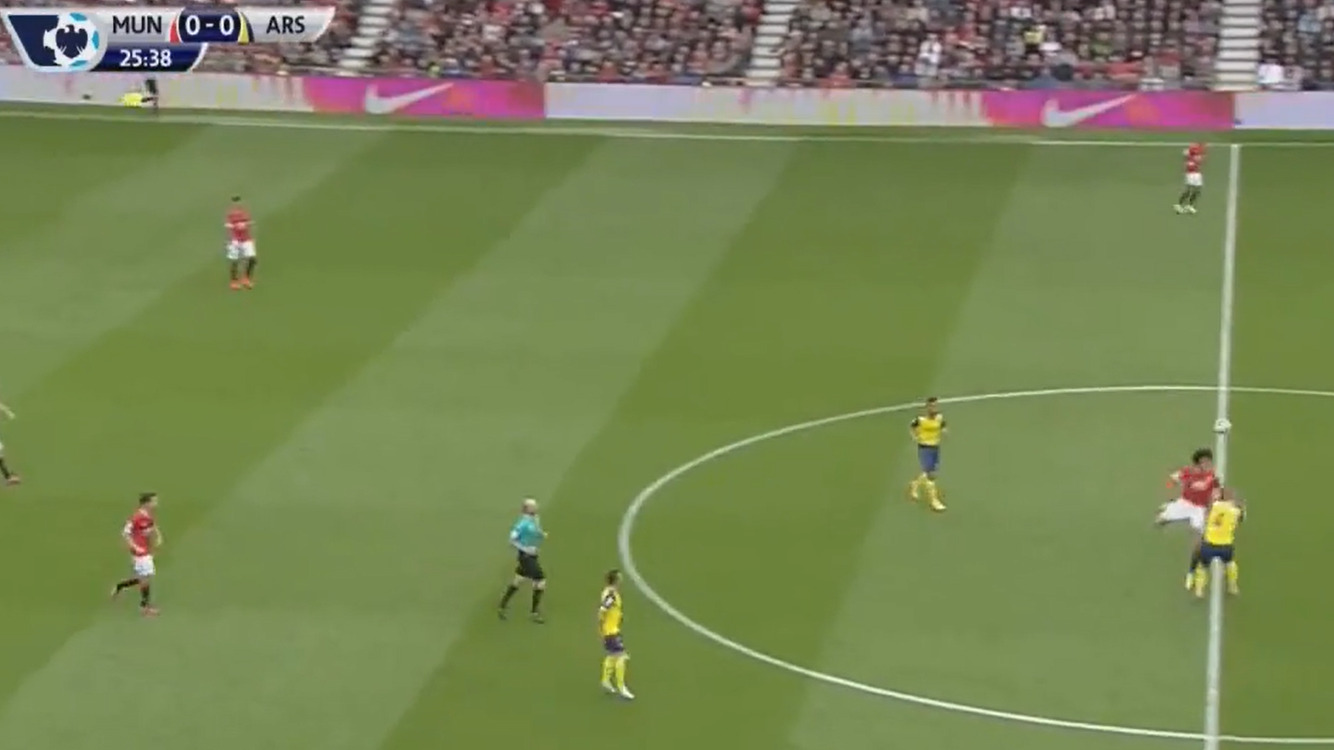}
    \end{minipage}\hfill
    \begin{minipage}{0.50\textwidth}
        \caption{An illustrative failure: VARS misclassifies an \textit{elbowing} as a
        \textit{challenge}, removing relevant precedents before semantic ranking.}
        \label{fig:qualitative_failure}
    \end{minipage}
\end{figure}

\Cref{fig:qualitative_failure} shows an illustrative failure of the text-mediated
pipeline. The query is an \textit{elbowing} offence that VARS misclassifies as a
\textit{challenge}. Because the VARS categorical filter enforces strict category
equality, valid \textit{elbowing} precedents are removed before semantic ranking
and cannot be recovered by later stages. This example highlights a limitation of
the VARS filtering stage: an early classification error can eliminate otherwise
relevant precedents.

\section{Conclusion}
\label{sec:conclusion}

We introduced \textbf{SoccerNet-FoulRet}, the first benchmark for semantic foul
retrieval, framing refereeing consistency as a video-to-video retrieval problem.
Rather than automating the referee's verdict, the goal is to retrieve relevant
historical precedents while leaving the final decision to the referee. We evaluate
this task using human-verified precedent judgments and category-relevance labels.
Our experiments show that semantic foul retrieval remains challenging. Intern-Video2-1B
is the strongest zero-shot retriever overall, yet achieves under $5\%$ HitRate@10
on the human-verified precedent evaluation. Performance also degrades sharply as
category relevance tightens from Action to Action + Offence + Severity, showing that current models are better at matching broad foul categories than distinguishing the finer details that affect refereeing decisions. Category-supervised fine-tuning improves category relevance substantially but shows only modest transfer to the human-verified precedent evaluation, further showing that the type of action, offence, and severity of a foul does not fully capture precedent relevance.
Several limitations remain. First, the human-verified relevance pool is
incomplete. Only a small number of candidates are judged per query, so unjudged
retrieved clips may themselves be valid precedents. Second, the precedent judgments come from a limited annotator pool and therefore do not establish broader agreement across referees.
These limitations motivate denser precedent annotations and retrieval models that
better capture refereeing semantics such as intent, point of contact, and severity. We hope SoccerNet-FoulRet provides a foundation for
retrieval systems that can ultimately support more consistent football refereeing.

\noindent\textbf{Acknowledgement.}
The research reported in this publication was supported by funding from KAUST CoE for Generative AI, under award number 5940.

% \section{Camera-Ready Manuscript Preparation}
% \label{sec:manuscript}
% This information will follow after paper decisions have been announced.

% \clearpage\mbox{}Page \thepage\ of the manuscript.
% \clearpage\mbox{}Page \thepage\ of the manuscript.
% \clearpage\mbox{}Page \thepage\ of the manuscript.
% \clearpage\mbox{}Page \thepage\ of the manuscript.
% \clearpage\mbox{}Page \thepage\ of the manuscript. This is the last page.
% Now we have reached the maximum length of an ACCV \ACCVyear{} submission (excluding references).
% References should start immediately after the main text, but can continue past p.\ 14 if needed.

\bibliographystyle{splncs04-MVD}
\bibliography{bib/abbreviation-short,bib/main}

@STRING{arxiv = "CoRR"}

@STRING{arxiv = "arXiv"}

@STRING{acm-c-aicm = "ACM Int. Conf. Multimedia"}

@STRING{acm-c-sigir = "ACM Int. Conf. Res. Dev. Inf. Retr. (SIGIR)"}

@STRING{publisher-ieee = "Inst. Electr. Electron. Eng. (IEEE)"}

@STRING{publisher-ieee-short = "IEEE"}

@STRING{ieee-tm = "IEEE Trans. Multimedia"}

@STRING{ieee-c-cvpr = "IEEE Conf. Comput. Vis. Pattern Recognit. (CVPR)"}

@STRING{ieee-c-cvsports = "IEEE Int. Conf. Comput. Vis. Pattern Recognit. Work. (CVPRW), CVsports"}

@STRING{ieee-c-iccvx = "IEEE/CVF Int. Conf. Comput. Vis. (ICCV)"}

@STRING{ieee-c-icvprx = "IEEE/CVF Conf. Comput. Vis. Pattern Recognit. (CVPR)"}

@STRING{ieee-c-icvprwx = "IEEE/CVF Conf. Comput. Vis. Pattern Recognit. Work. (CVPRW)"}

@STRING{publisher-springerip = "Springer Int. Publ."}

@STRING{publisher-springersbm = "Springer Sci. Bus. Media LLC"}

@STRING{publisher-springerswiss = "Springer Nat. Switz."}

@STRING{as = "Acta Ster."}

@STRING{scid = "Sci. Data"}

@STRING{lncs   = "Lect. Notes Comput. Sci."}

@STRING{eccv = "Eur. Conf. Comput. Vis. (ECCV)"}

@STRING{iclr = "Int. Conf. Learn. Represent. (ICLR)"}

@STRING{nips = "Adv. Neural Inf. Process. Syst. (NeurIPS)"}

@STRING{publisher-curran = "Curran Assoc. Inc."}

@STRING{city-nashville = "Nashville, TN, USA"}

@STRING{city-neworleans = "New Orleans, LA, USA"}

@STRING{city-paris = "Paris, Fr."}

@STRING{city-portland = "Portland, OR, USA"}

@STRING{city-saltlake = "Salt Lake City, UT, USA"}

@STRING{city-sandiego = "San Diego, CA, USA"}

@STRING{city-seattle = "Seattle, WA, USA"}

@STRING{city-seoul = "Seoul, South Korea"}

@STRING{city-singapore = "Singapore"}

@STRING{city-vancouver = "Vancouver, Can."}

@STRING{city-washington = "Washington, DC, USA"}

@STRING{Jan = "Jan."}

@STRING{Apr = "Apr."}

@STRING{Jun = "Jun."}

@STRING{Jul = "Jul."}

@STRING{Sep = "Sept."}

@STRING{Oct = "Oct."}

@STRING{Nov = "Nov."}

@STRING{Dec = "Dec."}

@inproceedings{Jiang2014VCDB,
	title = {{VCDB}: A Large-Scale Database for Partial Copy Detection in Videos},
	author = {Jiang, Yu-Gang and Jiang, Yudong and Wang, Jiajun},
	booktitle = eccv,
	series = lncs, 
	volume = {8692},
	pages = {357-371},
	year = {2014},
	publisher = publisher-springerip,
	doi = {10.1007/978-3-319-10593-2_24},
	url = {https://doi.org/10.1007/978-3-319-10593-2_24}
}

@inproceedings{Jiang2019SVD,
	title = {{SVD}: A Large-Scale Short Video Dataset for Near-Duplicate Video Retrieval},
	author = {Jiang, Qing-Yuan and He, Yi and Li, Gen and Lin, Jian and Li, Lei and Li, Wu-Jun},
	booktitle = ieee-c-iccvx,
	pages = {5280-5288},
	month = Oct,
	year = {2019},
	publisher = publisher-ieee-short,
	address = city-seoul,
	d2b-version = {1.0},
	doi = {10.1109/iccv.2019.00538},
	url = {https://doi.org/10.1109/iccv.2019.00538}
}

@inproceedings{Giancola2018SoccerNet,
	title = {{SoccerNet}: A Scalable Dataset for Action Spotting in Soccer Videos},
	author = {Giancola, Silvio and Amine, Mohieddine and Dghaily, Tarek and Ghanem, Bernard},
	booktitle = ieee-c-icvprwx,
	pages = {1792-179210},
	month = Jun,
	year = {2018},
	publisher = publisher-ieee-short,
	address = city-saltlake,
	doi = {10.1109/cvprw.2018.00223},
	url = {https://doi.org/10.1109/cvprw.2018.00223}
}

@inproceedings{Deliege2021SoccerNetv2,
	title = {{SoccerNet}-v2: A Dataset and Benchmarks for Holistic Understanding of Broadcast Soccer Videos},
	author = {Deli{\`e}ge, Adrien and Cioppa, Anthony and Giancola, Silvio and Seikavandi, Meisam J. and Dueholm, Jacob V. and Nasrollahi, Kamal and Ghanem, Bernard and Moeslund, Thomas B. and Van Droogenbroeck, Marc},
	booktitle = ieee-c-icvprwx,
	pages = {4503-4514},
	month = Jun,
	year = {2021},
	publisher = publisher-ieee-short,
	address = city-nashville,
	doi = {10.1109/cvprw53098.2021.00508},
	url = {https://doi.org/10.1109/CVPRW53098.2021.00508}
}

@inproceedings{Cioppa2022SoccerNetTracking,
	title = {{SoccerNet}-Tracking: Multiple Object Tracking Dataset and Benchmark in Soccer Videos},
	author = {Cioppa, Anthony and Giancola, Silvio and Deli\`ege, Adrien and Kang, Le and Zhou, Xin and Cheng, Zhiyu and Ghanem, Bernard and Van Droogenbroeck, Marc},
	booktitle = ieee-c-cvsports,
	pages = {3490-3501},
	month = Jun,
	year = {2022},
	publisher = publisher-ieee-short,
	address = city-neworleans,
	doi = {10.1109/cvprw56347.2022.00393},
	url = {https://doi.org/10.1109/CVPRW56347.2022.00393}
}

@article{Cioppa2022Scaling,
	title = {Scaling up {SoccerNet} with multi-view spatial localization and re-identification},
	author = {Cioppa, Anthony and Deli{\`e}ge, Adrien and Giancola, Silvio and Ghanem, Bernard and Van Droogenbroeck, Marc},
	journal = scid,
	volume = {9},
	number = {1},
	pages = {1-9},
	numpages = {9},
	month = Jun,
	year = {2022},
	publisher = publisher-springersbm,
	doi = {10.1038/s41597-022-01469-1},
	url = {https://doi.org/10.1038/s41597-022-01469-1}
}

@inproceedings{Mkhallati2023SoccerNetCaption,
	title = {{SoccerNet}-Caption: Dense Video Captioning for Soccer Broadcasts Commentaries},
	author = {Mkhallati, Hassan and Cioppa, Anthony and Giancola, Silvio and Ghanem, Bernard and Van Droogenbroeck, Marc},
	booktitle = ieee-c-icvprwx,
	pages = {5074-5085},
	month = Jun,
	year = {2023},
	publisher = publisher-ieee-short,
	address = city-vancouver,
	doi = {10.1109/cvprw59228.2023.00536},
	url = {https://doi.org/10.1109/CVPRW59228.2023.00536}
}

@inproceedings{Held2023VARS,
	title = {{VARS}: Video Assistant Referee System for Automated Soccer Decision Making from Multiple Views},
	author = {Held, Jan and Cioppa, Anthony and Giancola, Silvio and Hamdi, Abdullah and Ghanem, Bernard and Van Droogenbroeck, Marc},
	booktitle = ieee-c-icvprwx,
	pages = {5086-5097},
	month = Jun,
	year = {2023},
	publisher = publisher-ieee-short,
	address = city-vancouver,
	doi = {10.1109/cvprw59228.2023.00537},
	url = {https://doi.org/10.1109/CVPRW59228.2023.00537}
}

@inproceedings{Held2024XVARS,
	title = {{X-VARS}: Introducing Explainability in Football Refereeing with Multi-Modal Large Language Models},
	author = {Held, Jan and Itani, Hani and Cioppa, Anthony and Giancola, Silvio and Ghanem, Bernard and Van Droogenbroeck, Marc},
	booktitle = ieee-c-icvprwx,
	pages = {3267-3279},
	month = Jun,
	year = {2024},
	publisher = publisher-ieee-short,
        address = city-seattle,
	doi = {10.1109/cvprw63382.2024.00332},
	url = {https://doi.org/10.1109/CVPRW63382.2024.00332}
}

@inproceedings{Somers2024SoccerNetGameState,
	title = {{SoccerNet} Game State Reconstruction: End-to-End Athlete Tracking and Identification on a Minimap},
	author = {Somers, Vladimir and Joos, Victor and Cioppa, Anthony and Giancola, Silvio and Ghasemzadeh, Seyed Abolfazl and Magera, Floriane and Standaert, Baptiste and Mansourian, Amir M. and Zhou, Xin and Kasaei, Shohreh and Ghanem, Bernard and Alahi, Alexandre and Van Droogenbroeck, Marc and De Vleeschouwer, Christophe},
	booktitle = ieee-c-icvprwx,
	pages = {3293-3305},
	month = Jun,
	year = {2024},
	publisher = publisher-ieee-short,
        address = city-seattle,
	doi = {10.1109/cvprw63382.2024.00334},
	url = {https://doi.org/10.1109/CVPRW63382.2024.00334}
}

@article{Li2026Qwen3VLEmbedding-arxiv,
	title = {Qwen3-{VL}-Embedding and Qwen3-{VL}-Reranker: A Unified Framework for State-of-the-Art Multimodal Retrieval and Ranking},
	author = {Li, Mingxin and Zhang, Yanzhao and Long, Dingkun and Chen, Keqin and Song, Sibo and Bai, Shuai and Yang, Zhibo and Xie, Pengjun and Yang, An and Liu, Dayiheng and Zhou, Jingren and Lin, Junyang},
	journal = arxiv,
	volume = {abs/2601.04720},
	year = {2026},
	publisher = arxiv,
	eprint = {2601.04720},
	eprinttype = {arXiv},
	doi = {10.48550/arXiv.2601.04720},
	url = {https://doi.org/10.48550/arXiv.2601.04720}
}

@inproceedings{Kraaij2011TRECVID,
	title = {{TRECVID} 2011 Content-Based Copy Detection: Task Overview},
	author = {Kraaij, Wessel and Awad, George},
	booktitle = {Proceedings of the {TRECVID} 2011 Workshop},
	pages = {1-36},
	month = Dec,
	year = {2011},
	address = {Gaithersburg, MD, USA}
}

@article{KordopatisZilos2019FIVR,
	title = {{FIVR}: Fine-Grained Incident Video Retrieval},
	author = {Kordopatis-Zilos, Giorgos and Papadopoulos, Symeon and Patras, Ioannis and Kompatsiaris, Ioannis},
	journal = ieee-tm,
	volume = {21},
	number = {10},
	pages = {2638-2652},
	month = Oct,
	year = {2019},
	publisher = publisher-ieee,
	d2b-version = {1.0},
	doi = {10.1109/tmm.2019.2905741},
	url = {https://doi.org/10.1109/TMM.2019.2905741}
}

@inproceedings{Wray2021OnSemantic,
	title = {On Semantic Similarity in Video Retrieval},
	author = {Wray, Michael and Doughty, Hazel and Damen, Dima},
	booktitle = ieee-c-icvprx,
	pages = {3649-3659},
	month = Jun,
	year = {2021},
	publisher = publisher-ieee-short,
	address = city-nashville,
	doi = {10.1109/cvpr46437.2021.00365},
	url = {https://doi.org/10.1109/CVPR46437.2021.00365}
}

@inproceedings{Xiao2024CPack,
	title = {{C-Pack}: Packed Resources For General {Chinese} Embeddings},
	author = {Xiao, Shitao and Liu, Zheng and Zhang, Peitian and Muennighoff, Niklas and Lian, Defu and Nie, Jian-Yun},
	booktitle = acm-c-sigir,
	pages = {641-649},
	month = Jul,
	year = {2024},
	publisher = {ACM},
	address = city-washington,
	d2b-version = {1.0},
	doi = {10.1145/3626772.3657878},
	url = {https://doi.org/10.1145/3626772.3657878}
}

@inproceedings{Revaud2013Event,
	title = {Event Retrieval in Large Video Collections with Circulant Temporal Encoding},
	author = {Revaud, Jerome and Douze, Matthijs and Schmid, Cordelia and Jegou, Herve},
	booktitle = ieee-c-cvpr,
	pages = {2459-2466},
	month = Jun,
	year = {2013},
	publisher = publisher-ieee-short,
	address = city-portland,
	doi = {10.1109/cvpr.2013.318},
	url = {https://doi.org/10.1109/CVPR.2013.318}
}

@inproceedings{Song2011Multiple,
	title = {Multiple feature hashing for real-time large scale near-duplicate video retrieval},
	author = {Song, Jingkuan and Yang, Yi and Huang, Zi and Shen, Heng Tao and Hong, Richang},
	booktitle = acm-c-aicm,
	pages = {423-432},
	month = Nov,
	year = {2011},
	publisher = {ACM},
	address = {Scottsdale, Arizona, USA},
	doi = {10.1145/2072298.2072354},
	url = {https://doi.org/10.1145/2072298.2072354}
}

@inproceedings{Wu2007Practical,
	title = {Practical elimination of near-duplicates from web video search},
	author = {Wu, Xiao and Hauptmann, Alexander G. and Ngo, Chong-Wah},
	booktitle = acm-c-aicm,
	pages = {218-227},
	month = Sep,
	year = {2007},
	publisher = {ACM},
	address = {Augsburg Germany},
	doi = {10.1145/1291233.1291280},
	url = {https://doi.org/10.1145/1291233.1291280}
}

@inproceedings{Cui2023SportsMOT,
	title = {{SportsMOT}: A Large Multi-Object Tracking Dataset in Multiple Sports Scenes},
	author = {Cui, Yutao and Zeng, Chenkai and Zhao, Xiaoyu and Yang, Yichun and Wu, Gangshan and Wang, Limin},
	booktitle = ieee-c-iccvx,
	pages = {9887-9897},
	month = Oct,
	year = {2023},
	publisher = publisher-ieee-short,
	address = city-paris,
	doi = {10.1109/iccv51070.2023.00910},
	url = {https://doi.org/10.1109/ICCV51070.2023.00910}
}

@inproceedings{Liberatori2025ConViSBench,
	title = {{ConViS}-Bench: Estimating Video Similarity Through Semantic Concepts},
	author = {Liberatori, Benedetta and Conti, Alessandro and Vaquero, Lorenzo and Wang, Yiming and Ricci, Elisa and Rota, Paolo},
        booktitle = nips,
	volume = {39},
        pages = {1-32},
        month = Dec,
        year = {2025},
        publisher = publisher-curran,
        address = city-sandiego,
        url = {https://openreview.net/forum?id=NoIWLerNKH}
}

@inproceedings{Jiang2025VLM2Vec,
	title = {{VLM2Vec}: Training Vision-Language Models for Massive Multimodal Embedding Tasks},
	author = {Jiang, Ziyan and Meng, Rui and Yang, Xinyi and Yavuz, Semih and Zhou, Yingbo and Chen, Wenhu},
	booktitle = iclr,
	pages = {1-25},
	month = Apr,
	year = {2025},
	address = city-singapore,
	url = {https://openreview.net/forum?id=TE0KOzWYAF}
}

@inproceedings{Liu2025LamRA,
	title = {{LamRA}: Large Multimodal Model as Your Advanced Retrieval Assistant},
	author = {Liu, Yikun and Zhang, Yajie and Cai, Jiayin and Jiang, Xiaolong and Hu, Yao and Yao, Jiangchao and Wang, Yanfeng and Xie, Weidi},
	booktitle = ieee-c-icvprx,
	pages = {4015-4025},
	month = Jun,
	year = {2025},
	publisher = publisher-ieee-short,
	address = city-nashville,
	doi = {10.1109/cvpr52734.2025.00380},
	url = {https://doi.org/10.1109/CVPR52734.2025.00380}
}

@inproceedings{Wang2024InternVideo2,
	title = {{InternVideo2}: Scaling Foundation Models for Multimodal Video Understanding},
	author = {Wang, Yi and Li, Kunchang and Li, Xinhao and Yu, Jiashuo and He, Yinan and Chen, Guo and Pei, Baoqi and Zheng, Rongkun and Wang, Zun and Shi, Yansong and Jiang, Tianxiang and Li, Songze and Xu, Jilan and Zhang, Hongjie and Huang, Yifei and Qiao, Yu and Wang, Yali and Wang, Limin},
	booktitle = eccv,
	series = lncs,
	volume = {15143},
	pages = {396-416},
	month = Nov,
	year = {2024},
	publisher = publisher-springerswiss,
	doi = {10.1007/978-3-031-73013-9_23},
	url = {https://doi.org/10.1007/978-3-031-73013-9_23}
}

@inproceedings{Buckley2004Incomplete,
  author    = {Chris Buckley and Ellen M. Voorhees},
  title     = {Retrieval Evaluation with Incomplete Information},
  booktitle = {Proceedings of the 27th Annual International ACM SIGIR
               Conference on Research and Development in Information Retrieval},
  pages     = {25--32},
  year      = {2004},
  doi       = {10.1145/1008992.1009000}
}

\clearpage
\appendix

\begin{center}
    {\LARGE\bfseries
    SoccerNet-FoulRet: Retrieving Semantically Similar Soccer Foul Videos
    \par}

    \vspace{0.6em}

    {\Large\bfseries Supplementary Material\par}
\end{center}

\vspace{1.2em}

\section{Task-Specific Fine-Tuning}
\label{sec:supp_finetune}
\noindent\textbf{Fine-tuning objective.}
We fine-tune Qwen3-VL-Embedding-2B using supervised contrastive learning at three
levels of similarity: Action, Action+Offence, and Action+Offence+Severity. At each
level, different clips with matching labels are treated as positives, while clips
with non-matching labels provide contrastive negatives. Self-pairs are excluded.

For an anchor $i$ at level $t$, let $P_i^{(t)}$ be the set of other clips with matching labels and $C_i^{(t)}$ the set of all eligible non-self clips in the minibatch. Thus, $P_i^{(t)}\subseteq C_i^{(t)}$. The query-to-gallery loss is
\[
\ell_i^{(t)}
=
-\log
\frac{
    \sum_{j\in P_i^{(t)}}
    \exp\!\left(q_i^\top g_j/\tau\right)
}{
    \sum_{j\in C_i^{(t)}}
    \exp\!\left(q_i^\top g_j/\tau\right)
},
\]
where $q_i$ and $g_j$ are $\ell_2$-normalized query and gallery embeddings and $\tau=0.05$. We also compute the corresponding loss in the reverse gallery-to-query direction, average the two directions, and then average the three label-level losses equally. Clips without severity annotations contribute to the Action and Action+Offence losses but are omitted from the Action+Offence+Severity loss. Training uses label-balanced batches of 12 clips, comprising six label groups with two distinct clips per group. For each anchor, the other clip in its group provides one guaranteed positive, while the clips from the remaining groups provide the candidate negatives at the strictest hierarchy level. At the coarser levels, clips from different groups become additional positives whenever they share the same Action or Action+Offence labels.

\noindent\textbf{Implementation details.}
We train for 10 epochs using AdamW with a learning rate of $10^{-5}$ and weight
decay of $0.05$, applied only to the trainable LoRA parameters. We use a cosine
learning-rate schedule with 10\% warmup, gradient accumulation over eight steps,
and gradient clipping at a maximum norm of $1.0$. LoRA is configured with rank
$16$, scaling factor $\alpha=16$, and dropout $0.10$. All experiments use a fixed
random seed of $42$.

\section{Oracle Candidate Generation}
\label{sec:supp_candidate}
To avoid auditing 2M+ pairs, candidates are filtered using SoccerNet-MVFoul labels (Action Class, Offence, Contact, Body Part) and ranked by SBERT similarity over ground-truth referee-written XVARS \cite{Held2024XVARS} explanations ($\sim$70 professional referees), averaging cosine similarities across three jurisprudential Q\&A tracks (covering physical mechanics, rule interpretation, disciplinary severity) to preserve diversity. This reduces annotation overhead by $99.83\%$ while concentrating review on
plausible precedent candidates.

\section{Relevance Structure}
\label{sec:supp_relevance}
The benchmark natively features a query-conditioned, many-to-many relevance structure where multiple queries can share precedents, accurately reflecting sports jurisprudence. Each query--candidate pair was judged independently. Across the 693 evaluated queries, 2,751 approved pairs span 1,353 unique precedents. Of these 1,353 precedents, 691 precedents match multiple queries, and 652 queries share at least one precedent.

\section{Annotation Protocol}
\label{sec:supp_annotation}
Pairs were labeled \textit{Valid} only if matching both "Form" (physical mechanics) and "Substance" (regulatory severity/tactical context), going beyond the four-attribute coarse filter (Action Class, Offence, Contact, Body Part): pitch location matters, e.g. an identical tackle differs in midfield versus inside the box denying a clear goal-scoring opportunity \ref{fig:similar_but_different}. A single annotator with 18 years of soccer experience and IFAB knowledge annotated the 3,518 approved/rejected pairs ($\sim$50 hours).

\section{Inter-Annotator Agreement}
\label{sec:supp_agreement}

To assess annotation consistency, a second annotator (one of the authors of this paper) independently reviewed a subset of 101 queries, corresponding to 502 query--candidate pairs. Of these, 99 queries have five judged candidates, while two have annotation pools of four and three candidates, respectively. Across these pairs, raw agreement was $67.13\%$, with Cohen's $\kappa=0.238$.

Agreement was higher for accepted precedents ($76.46\%$) than for rejected candidates ($45.54\%$). These results indicate that annotators agree more consistently on accepted precedents. This reflects the subjective nature of precedent relevance and motivates broader annotation with multiple refereeing perspectives in future work.
\end{document}